\documentclass[sigconf, nonacm]{acmart}
\AtBeginDocument{%
  }

\usepackage{booktabs}
\usepackage{threeparttable}
\usepackage{subcaption}
\usepackage{rotating}
\usepackage{algorithm}       
\usepackage{algorithmic}
\usepackage{mathtools} 
\usepackage{amsmath}

\newcommand{\PARAMETERS}[1]{\item[\textbf{Parameters:}] #1}

\begin{document}

\title{Temporal Knowledge Graph Hyperedge Forecasting: Exploring Entity-to-Category Link Prediction}

\author{Edward Markai}
\affiliation{%
  \institution{KTH Royal Institute of Technology}
  \city{Stockholm}
  \country{Sweden}
}
\email{markai@kth.se}

\author{Sina Molavipour}
\affiliation{%
  \institution{SEB Group}
  \city{Stockholm}
  \country{Sweden}}
\email{sina.molavipour@seb.se}


\begin{abstract}
Temporal Knowledge Graphs have emerged as a powerful way of not only modeling static relationships between entities but also the dynamics of how relations evolve over time. As these informational structures can be used to store information from a real-world setting, such as a news flow, predicting future graph components to a certain extent equates predicting real-world events. Most of the research in this field focuses on embedding-based methods, often leveraging convolutional neural net architectures. These solutions act as black boxes, limiting insight. In this paper, we explore an extension to an established rule-based framework, TLogic, that yields a high accuracy in combination with explainable predictions. This offers transparency and allows the end-user to critically evaluate the rules applied at the end of the prediction stage. The new rule format incorporates entity category as a key component with the purpose of limiting rule application only to relevant entities. When categories are unknown for building the graph, we propose a data-driven method to generate them with an LLM-based approach. Additionally, we investigate the choice of aggregation method for scores of retrieved entities when performing category prediction. 

\end{abstract}


\maketitle

\newcommand\blfootnote[1]{%
  \begingroup
  \renewcommand\thefootnote{}\footnote{#1}%
  \addtocounter{footnote}{-1}%
  \endgroup
}

\blfootnote{\footnotesize AI for Finance Symposium part of ACM ICAIF '25 Conference, Singapore, November 2025}

\section{Introduction}
\label{sec:intro}
Datasets in application areas such as financial information, international politics, or patient health records are deeply interconnected. In addition, they do not remain static, but in reality evolve over time. This temporal aspect poses many different fundamental challenges. Some examples cover reconciliation of conflicting information, accurately describing the evolution of events over time. Traditional data structures are often not able to represent and query relationships at high complexity. Temporal Knowledge Graphs (TKGs) address these challenges by embedding temporal information directly into the graph structure (each node (entity) and edge (relation) have a timestamp). The result is a versatile foundation for temporal reasoning and event prediction \cite{trivedi2017know}.

More formally, Knowledge Graphs (KGs) structures data in a graph format, with nodes and often directed edges. The static data structures consist of facts or events represented as triples, where two nodes (subject and object which are both entities) are connected via an edge (relation) describing the association between the two as \textit{(subject, relation, object)}.
An example is \textit{(The US president, criticize, European Union)}. Combining a large number of facts in one graph forms a comprehensive summary of information, often stemming from a textual knowledge source such as a news flow. To incorporate temporal representation, TKGs are introduced to describe how information dynamically evolves over time by adding a temporal dimension to each fact or event \textit{(subject, relation, object, timestamp)}
The previously mentioned example can now be described at a certain point in time as a quadruple \textit{(The US president, criticize, European Union, 15th Dec 2016)}. 

TKGs show promising results in forecasting real-world events \cite{renet, re-gcn, zrLLM}. In this paper, we focus on link prediction, the task of predicting the object of a quadruple at a future timestamp given a subject and a relation, such as \textit{(subject, relation, ?, future timestamp)}. Moreover, we explore hyperedge prediction, which we define as the link between a subject and a category, encompassing all the relations between that subject and the objects of the given category. Our method is an extension of an established rule-based framework, TLogic \cite{tlogic}. TLogic is a rule-based method that predict future events by identifying sequential patterns in data that satisfy a predefined format and that historically have shown to hold some predictive power.
For example, if the event \textit{(Entity 1, announce partnership with, Entity 2, $T_2$)} historically in 54\% of cases has been followed by the event \textit{(Entity 2, acquires, Entity 1, $T_1$)}, a rule can be derived to capture this pattern. The process of identifying such patterns and formalizing them into rules is known as rule mining or training \cite{tlogic}. Note that in this context, \textit{Entity 1} and \textit{Entity 2} serve as placeholders for the same entities across both facts, and $T_1$ and $T_2$ represent placeholders for timestamps to which some condition applies, in this case $T_2>T_1$. In practice, if the event \textit{(Nuance, announce partnership, Microsoft, Feb 2019)} occur in a TKG, the rule described above may be leveraged to predict the upcoming event \textit{(Microsoft, acquires, Nuance, April 2021)}. This application of a rule to concrete datapoints, such as assigning \textit{Nuance} to \textit{Entity 1} and so forth, is referred to as rule grounding \cite{tlogic}. 
As incorporating categories would provide more descriptive power to the rules, we extend the rule format of TLogic into a new format which we denote as C-TLogic:
\begin{equation}
    \label{eq:sextuple}
    \textit{(subject, relation, object, timestamp, subject type, object type).}
\end{equation}
By integrating these in the rule format and mining process, the aim is to limit rule application only to those facts with entities where the rules are relevant. In the rest of the paper, "category" and "type" are used interchangeably. 

In Section~\ref{sec:problem} interpretable TKG forecasting is motivated and C-TLogic is introduced. Next, a systematic method to assign categories is presented, and finally we provide formal definitions for rule learning and application based on TLogic \cite{tlogic}. In Section~\ref{sec:implementation}, the components of the learner and retriever are studied in detail. Additionally we propose how the model should be evaluated considering the category prediction task. The performance of different frameworks is compared and the results are provided in~Section~\ref{sec:res_and_an} for both TLogic and C-TLogic. Finally, we analyze the results and conclude the paper.

\section{Problem}
\label{sec:problem}

The large majority of work in predicting the evolution of TKGs relies on models that learn embeddings for graph constituents and utilize convolutional neural networks for prediction, such as in \cite{renet, re-gcn, zrLLM}. However, these approaches lack explicit interpretability, making it difficult to qualitatively evaluate predictions. To address this, rule-based methods that use logically motivated rule formats have been proposed. These methods often apply rules en masse and aggregate the predictions into a sorted ranking that comprises the final results.

This paper aims to extend the work of TLogic, an established rule-based method for TKG link prediction \cite{tlogic}. TLogic generates strong predictive results compared to other methods. As it lacks a mechanism for facilitating rule learning and rule grounding based on entity type, a rule mined from entities of certain types can unfavorably be applied to sequences of facts involving an entirely different set of entity types, where the rule is irrelevant. For example, a rule derived from data on the actions of an NGO is likely irrelevant to companies, yet may still be applied in such contexts.

Furthermore, in many scenarios, it may be more valuable to predict the category of an object in a future fact rather than the specific object itself. Here, we distinguish between entity-predictions where the task is to predict the entity of a future fact and category-specific predictions that aim to predict which category an entity will belong to at a future timestamp. In entity-prediction, the TKG data may not contain sufficient information to support a valuable entity-specific prediction. In addition, only focusing on the top-ranked entity-specific predictions may overlook important general insights when instead considering all predictions within each category (see Section \ref{sec:rule_application} for how this is applied in practice). This may also make the model less sensitive to noise. As an example, consider the task of predicting what might negatively impact a company at a future timestamp, expressed as the query
\begin{equation}
    \label{eq:sextuple_query}
    \textit{(Company A, negatively impacted by, ?, t, Tech Company, ?)},
\end{equation}
based on the format~\eqref{eq:sextuple}. While the top-ranked predictions suggest a range of specific entities, all predictions produced by the rules may collectively indicate some overarching pattern or insight. Suppose that a large number of entity-specific predictions, such as \textit{Inflation}, \textit{CCI}, \textit{GDP} belong to the same category of \textit{Financial Indicator}. These entity predictions may not individually be ranked in the top, but collectively still indicate something important: that \textit{Company A} likely will be negatively affected by the approaching economic conditions.

Finally, by incorporating entity types into both the data and rule structures, additional information becomes available in the link forecasting query. As elaborated in Section~\ref{sec:rule_application}, relevant rules are extracted by matching the relation of the query to the relation in the fact that the rule predicts. Including the entity category in the query allows the alternative query formulation presented in~\eqref{eq:sextuple_query}. This can be further generalized by not fixing the relation but instead extracting the rules based on subject category. The same query in the more detailed format becomes 
\begin{equation}
    \label{eq:sextuple_query}
    \textit{(Company A, ?, ?, t, Tech Company, ?)},
\end{equation}
where the relation, object and object category are unknowns. This reformulation provides a broader contextual basis for prediction and essentially asks the question "What is next for \textit{Company A}?". This type of query may be more relevant for real-world application than limiting a query to a single subject-relation combination. While this last point is not part of the scope of this paper, it remains a relevant functionality that the format enables.

\subsection{Method}
%
Since the C-TLogic format extends TLogic by taking categories into account, it is required that categories of entities are provided in the datasets. Here, we begin with describing a systematic method to extract categories for each entity, and possibly extend them. Next, the notations of temporal logic rules for C-TLogic framework is introduced. Finally, we propose how to aggregate retrieved entities for the purpose of category prediction.

The purpose of the benchmark data used is the evaluation of the predictive framework that this work presents. In this field of research, it is common to rely on widely-used benchmark datasets, allowing for meaningful comparisons between different models. The datasets vary in context and methods of production. In this work, the ICEWS18 dataset and the FinDKG dataset was applied \cite{icews, findkg}. The ICEWS18 is a widely applied benchmark dataset and is the dataset on which the original TLogic framework is developed, while the FinDKG dataset is based on financial news and introduces a different structure. The ICEWS18 dataset contain events that can be described as autonomous actors against one another, such as:
\begin{equation*}
    \textit{(Angela Merkel, Consult, Head of Government (Sweden), t\footnotemark[1])},
\end{equation*}
while the FinDKG dataset with entity categories can describe more nuanced events and facts such as
\begin{equation*}
    \textit{(France, Has, Unrest, t\footnotemark[1], GPE, EVENT)}.
\end{equation*}

\footnotetext[1]{Note that the time has been exchanged for a $t$ as its nonessential for the example. See Section \ref{sec:intro} for explanation of the quadruple and sextuple format.}

To emphasize the point of this comparison, \textit{Unrest} is not an autonomous actor that could be included in the ICEWS18 dataset.
The motivation is to make the new framework comparable to the original framework while also applying the framework to the other dataset (FinDKG).

\subsection{Assigning categories}
\label{sec:assigning_categories}

To train type-aware rules on the non-categorized data such as the ICEWS18 dataset, entities have to be categorized as a pre-processing step. At the evaluation stage, both the type-aware rule format along with the categorization of the dataset is thus compared to the original TLogic format to conclude if the methods collectively add value in terms of prediction accuracy. In addition, expanding already existing categories into more granular sub-categories may be useful in practice. 


The ICEWS18 dataset does not include any entity categorization. For the same entities of different nationality, such as \textit{Citizen (Peru)}, nationality is specified. This does not apply to the entire dataset and post-production categorization in terms of nationality can thus not be applied for entities such as the \textit{World Bank}. To be able to compare the C-TLogic rules with the TLogic rules on the dataset on which the original method was developed, we categorize the entities based on the semantic similarity of the entity names.

To generate categories for ICEWS18 dataset, the text embeddings of the entity names are computed by OpenAI's text-embedding-3-large model and categories are found using Gaussian Mixture Model (GMM) clustering. GMM is a probabilistic model that can represent the distribution of a vector as a weighted sum of multiple Gaussian components to capture complex multi-modal structures. The meaning of a phrase can seldom be accurately classified as having only one interpretation in one context but instead includes a combination of several. 
To reduce dimensions of the embeddings before clustering, we use Principal Component Analysis (PCA).
A necessary task is choosing a reasonable number of clusters/categories. This is done using the Bayesian Information Criterion (BIC) and Akaike Information Criterion (AIC). In addition, K-means clustering is applied as a complement to validate the choice of the number of categories. 
By observing BIC and AIC across different number of PCA components, we choose 50 components and 100 clusters to categorize the entities of ICEWS18. Nevertheless, we process the dataset for 12 categories (with 50 PCA components) to have comparable results with FinDKG. Finally, to provide semantics for each category, we use OpenAI's o4-mini-high model to generate description for each category which can be found in Table \ref{tab:ICEWS18_12}. The generated datasets are denoted accordingly as ICEWS18\_100 and ICEWS18\_12.

\begin{figure}[htbp]
    \centering
    
    \begin{subfigure}{\linewidth}
    \centering
        \includegraphics[width=0.9\linewidth]{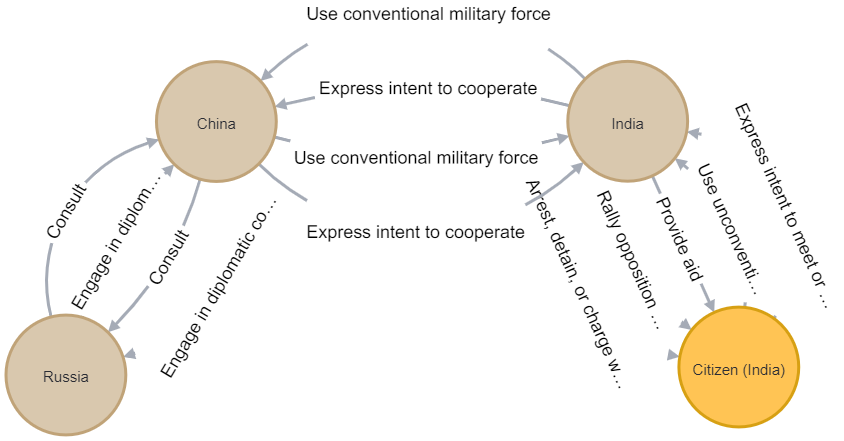}
        \label{fig:graph_kmeans3}
    \end{subfigure}

    \begin{subfigure}{\linewidth}
    \centering
        \includegraphics[width=0.9\linewidth]{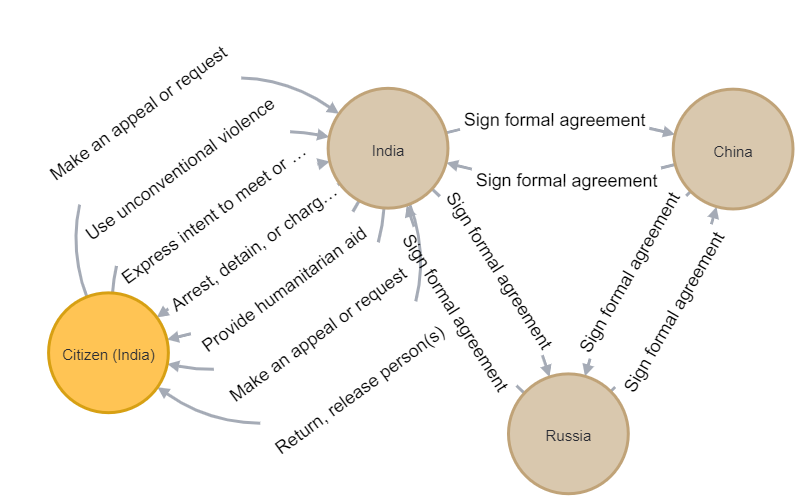}
        \label{fig:graph_kmeans4}
    \end{subfigure}
    
    \caption{Subset of top 10 entities by degree at timestamps 2018-11-04 (above), and 2018-11-06 (below) in the ICEWS18\_12 TKG. The color of the nodes indicate different entity categories.}
    \label{fig:comparison_graphs}
    \vspace{-15pt} 
\end{figure}


\begin{table*}[htbp]
    \centering
    {\small
        \begin{tabular}{@{}lll@{}}
        \toprule
        
        \textbf{Category} & \textbf{Definition} & \textbf{Example} \\ 
        \midrule
        $c_0$ & Individual people and political figures  & Vladimir Putin \\
        $c_1$ & Government ministries and departments & Treasury/Finance Ministry (Israel) \\
        $c_2$ & Government agencies and administrative bodies & Election Commission (India) \\
        $c_3$ & Legal and judiciary actors & Appeals Court (Egypt) \\
        $c_4$ & Religious organizations and clergy & Archdiocese (Australia) \\
        $c_5$ & Countries and territories & Kenya \\
        $c_6$ & Citizens and demographic groups & Citizen (India) \\
        $c_7$ & Educational and scholarly entities & Professor (Australia) \\
        $c_8$ & Legislators and parliamentary bodies & Member of Parliament (India) \\
        $c_9$ & Political parties and movements & Ruling Party (Nigeria) \\
        $c_{10}$ & Specific demographic segments & Women (Yemen) \\
        $c_{11}$ & Media organizations and outlets & Reuters \\ 
        \bottomrule
        \end{tabular}%
    }
    \caption{Resulting 12 entity categories based on GMM clustering of the ICEWS18 entity name embeddings.}
    \label{tab:ICEWS18_12}
\end{table*}

In FinDKG, categories are assigned by prompt-engineering a fine-tuned LLM \cite{findkg}. The model is instructed to extract entities from financial news articles and classify them into a predefined set of categories. The assignment is based on both contextual cues within the text and domain-specific rules that help disambiguate similar entities. The category assignments is made to enrich the resulting graph but also improve tasks such as link prediction. 

Much of the added value of category-specific prediction stem from downstream application. While predicting specific entities is valuable, a framework where type is predicted may prove complementary valuable in reality. This value increases with category granularity as the specificity of predictions increases. For this reason, we can expand the company category into subcategories using GICS sector as it is widely used in asset management and market research of financial institutes. The GICS taxonomy is a group of eleven categories (on the higher level) developed by MSCI and S\&P based on company industries into which companies can be classified worldwide by their principal business activity. An example for an application area is asset management where companies' actions in various scenarios and what they are affected by greatly differs across sectors. 

In addition to the cluster-based categorization described above, the categories of FinDKG are expanded. 
The taxonomy of GICS sectors are used to expand the category of \textit{COMP} into its subcategories. Due to the large number of entity members in the \textit{COMP} sector (13,645), the companies are classified using OpenAI's gpt-4o language model.  
The new dataset is referred to as FinDKG\_comp throughout the report. See Table \ref{tab:findkg_categories_comp} for a category overview of FinDKG\_comp.

\begin{table*}[htbp]
    \centering
    \small 
        \begin{tabular}{@{}lll@{}}
        \toprule
        \textbf{Category} & \textbf{Definition} & \textbf{Example} \\ 
        \midrule
        ORG     & Non-governmental and non-regulatory organisations     & World Bank \\
        ORG/GOV & Governmental bodies.                                  & U.S. Supreme Court \\
        ORG/REG & Regulatory bodies.                                    & Federal Reserve Board \\
        GPE     & Geopolitical entities like countries or cities.       & United Kingdom \\
        PERSON  & Individuals in influential or decision-making roles.  & Ursula von der Leyen \\
        COM\_SER\_COMP   & Companies of the communication services sector. & Verizon Communications Inc. \\
        CONS\_DISC\_COMP & Companies of the consumer discretionary sector. & Nike, Inc. \\
        CONS\_STAP\_COMP & Companies of the consumer staples sector. & Procter \& Gamble \\
        ENERGY\_COMP     & Companies of the energy sector. & Exxon Mobil \\
        FINANCIALS\_COMP & Companies of the financials sector. & JPMorgan Chase \\
        HEALTH\_CARE\_COMP & Companies of the health care sector. & Pfizer Inc. \\
        INDUSTRIALS\_COMP & Companies of the industrials sector. & General Electric Co. \\
        INFO\_TECH\_COMP & Companies of the information technology sector. & Nvidia Corp. \\
        MATERIALS\_COMP  & Companies of the materials sector. & U.S. Steel \\
        REAL\_ESTATE\_COMP & Companies of the real estate sector. & China Evergrande Group \\
        UTILITIES\_COMP  & Companies of the utilities sector. & Duke Energy \\
        PRODUCT & Tangible or intangible products or services. & Vaccine \\
        EVENT   & Material events with financial or economic implications. & Election Results \\
        SECTOR  & Sectors or industries in which companies operate.     & Technology Sector \\
        ECON\_INDICATOR & Non-numerical indicators of economic trends or states. & Mortgage rates \\
        FIN\_INSTRUMENT & Financial and market instruments.           & Nasdaq Composite Index \\
        CONCEPT & Abstract ideas, themes, or financial theories.        & Financial stability \\
        \bottomrule
        \end{tabular}
    \caption{Resulting entity categories from expanding the company (COMP) category of the FinDKG dataset into the GICS sector.}
    \label{tab:findkg_categories_comp}
\end{table*}

\subsection{Temporal Logic Rules}
\label{sec:tlogic_long}
This section presents the C-TLogic framework, extending the original TLogic \cite{tlogic}. While the notation has been adjusted to better fit the extension presented, much of the notations and definitions are equivalent to those of the original work. 

\subsubsection{Preliminaries}
\label{sec:prels}

Let $\mathcal{E}$ $\mathcal{R}$, $\mathcal{C}$ and $\mathcal{T}$ denote the set of entities (subjects and objects), relations, categories and timestamps in the TKG $\mathcal{G}$, respectively, such that $\mathcal{G} \subseteq \mathcal{E} \times  \mathcal{R} \times  \mathcal{E} \times \mathcal{T} \times \mathcal{C} \times \mathcal{C}$. The TKG consists of graphs at different timestamps $t \in \mathcal{T}$, where each node is an entity and each entity is a member of exactly one category (also referred to as "type") $c \in \mathcal{C}$. Every directed edge between entities are relations, $r \in \mathcal{R}$, in the direction from a subject $s \in \mathcal{E}$ to an object $o \in \mathcal{E}$. A fact in the TKG can thus be represented as a sextuple $(s,r,o,t, c^s, c^o)$. The entities are referred to as $s$ and $o$ to indicate subjects or objects but can also be referred to as $e$. The position of the variable in the tuple specifies the direction of the relation edge from the subject to the object. A hyper-edge refers to a relation from a subject entity to a category and exists as long as there is at least one relation from the subject to an object of the aforementioned category.

The link forecasting task has the goal of predicting the object of a fact, given the subject and relation at a future timestamp as $(s,r,?,t, c^s, ?)$. For every fact $(s,r,o,t, c^s, c^o) \in \mathcal{G}$, the fact with the inverse relation is denoted by {$(o,r^{-1},s,t, c^o, c^s)$}, with traversal of the edge in the opposite direction. By including these reverse edges, subject prediction is possible as $(o,r^{-1},?,t,  c_o, ?)$. The capital letters $S$, $O$, $E$ and $T$ denotes placeholders for actual entities from the sets $\mathcal{S}$, $\mathcal{O}$, $\mathcal{E}$ and $\mathcal{T}$, respectively. 


\subsubsection{Rule Learning}
\label{sec:rule_learning}

Here, the rule learning process is presented for C-TLogic. The purpose of a rule is to make a prediction of an entity in a future fact given a group of sequentially occurring facts that historically have shown to predate the predicted fact. The predicted fact is the rule head and the predictive sequence of facts is the rule body. The length of a rule equates the number of steps in the rule or the number of facts in the body. Grounding a rule means applying it to an actual sequence of facts in the data that satisfy the format of the rule. The rules are found by a random temporal walk $W$ through a TKG such that
%
%
%
\begin{equation}
    \label{eq:rand_walk}
    \begin{aligned}
        &W \coloneqq \\
        &\left\{ 
        \begin{array}{l}
            (s_{b+1}, r_{b+1}, o_{b+1}, t_{b+1}, c^s_{b+1}, c^o_{b+1}), \\ (s_{b}, r_{b}, o_{b}, t_{b}, c^s_{b}, c^o_{b}), \dots, \\
            (s_1, r_1, o_1, t_1, c^s_1, c^o_1)
        \end{array}
        :
        \begin{array}{l}
            t_{i+1} \geq t_i, \; o_{i+1}=s_i \\
            \forall i \in \{1, \dots, b\},  \\
            t_{b+1} > t_b, \; s_{b+1}=o_1
        \end{array}
        \right\}
    \end{aligned}
\end{equation}
where $b$ refers to the length of the rule.

Note that each object in an upcoming fact is equal to the subject in the current fact. In addition, the subject of the head fact equates the object of the first fact, satisfying cyclicality. The walk is made in reverse chronological order, as to why the indices in~\eqref{eq:rand_walk} incrementally increases to the left. Suppose we want to find a rule that predicts a fact with relation $r_{b+1}$. The first fact (the head, of index $b+1$) is sampled uniformly from the TKG $\mathcal{G}$ from all facts with relation $r_{b+1}$. 
The sampling of $v$ as the $m$:th fact in the walk from fact $u$ is made from the exponential distribution
\begin{equation}
    \label{eq:exp_distr}
    P(v; u, m) = {\exp(t_v - t_u)}/{\sum\nolimits_{\tilde{v} \in \mathcal{A}(m, u)} \exp(t_{\tilde{v}} - t_u)},
\end{equation}
where $m \in \{ b, b-1, \dots, 1 \}$ indicates which fact in the walk is sampled and $\mathcal{A}(m, u)$ denotes the sets of facts from which the $m$:th fact is sampled such that
\begin{equation*}
        \begin{aligned}
            &\mathcal{A}(m, u) :=\\
            &\begin{cases}
                \left\{(o_u, r, s_1, \hat{t}, c^{o}_u, c^{s}_1) \mid (o_u, r, s_1, \hat{t}, c^{o}_u, c^{s}_1) \in \tilde{\mathcal{G}},\ \hat{t} \leq t_u \right\} & m = 1\\
                \left\{(o_u, r, o, \hat{t}, c^{o}_u, c^o) \mid (o_u, r, o, \hat{t}, c^{o}_u, c^o) \in \mathcal{G},\ \hat{t} < t_u \right\} &  m = b, \\
                \left\{(o_u, r, o, \hat{t}, c^{o}_u, c^o) \mid (o_u, r, o, \hat{t}, c^{o}_u, c^o) \in \tilde{\mathcal{G}},\ \hat{t} \leq t_u \right\} & \text{otherwise},
            \end{cases}
        \end{aligned}
\end{equation*}
where \( \tilde{\mathcal{G}} := \mathcal{G} \setminus \left\{(o_u, r_u^{-1}, s_u, t_u, c^{o}_u, c^{s}_u) \right\} \) excludes the inverse edge to prevent walking back and forth over the same edge, and $s_1$ and $c^{s}_1$ indicates the subject and subject category of the first fact in the walk. Note that $\tilde{\mathcal{G}}$ depends on $u$.


A rule is created based on the sequence of facts in the temporal walk. Let a rule $R$ from a rulebank $\mathcal{B}$ be uniquely defined by the sequence of relations in the rule body and head, as well as the recurring order of entities (i.e. if the entity representing Ursula von der Leyen occurs twice in the walk, the placement of this recurring entity relative other entities is relevant and has to be remembered). Let the rule $R$ be of length $b$ and let the sequences of relations, subjects and objects from the random walk that the rule is based on be $\vec{r}=(r_{b+1}, r_{b}, r_{b-1}, \dots r_1)$, $\vec{s} = (s_{b+1}, s_{b}, s_{b-1}, \dots s_1)$ and $\vec{o} = (o_{b+1}, o_{b}, o_{b-1}, \dots o_1)$, respectively. While $\vec{s}$ and $\vec{o}$ are instantiations of entities from $\mathcal{G}$, $\vec{S}$ and $\vec{O}$ are vectors of placeholders to keep track of recurring entities' position within the rule. These vectors $\vec{r}$, $\vec{S}$, $\vec{O}$ in combination is unique for each rule. Let us define $R$ as:
\begin{equation}
\label{eq:second_rule}
    \begin{aligned}
       R \coloneqq \left\{ (S_i, r_i, O_i, T_i, c^s_i, c^o_i)_1^{b+1} : T_{j+1} \geq T_j \; \forall j \in \{1, \dots, b\}
        \right\}.
    \end{aligned}
\end{equation}
%
\noindent where $S_i$, $r_i$, $O_i$, $c^s_i$, $c^o_i$ denotes the $i$th element of $\vec{S}$, $\vec{r}$, $\vec{O}$, $\vec{c^S}$ and $\vec{c^O}$ respectively.
Note that the relation and category sequences of the rule remain the same as the walk $W$ while $\vec{S}$ and $\vec{O}$ act as placeholders to which we can assign actual entities when the rule later is applied. For a given rule $R$ we can find the rule body $B(R)$ and head $H(R)$ as sets of facts satisfying the relation sequence as
\begin{equation*}
    \begin{aligned}
        B(R) \coloneqq \Big\{ & (S_i, r_i, O_i, T_i, c^s_i, c^o_i)_{i=1}^b,
            \begin{array}{l}
            (S_i, r_i, O_i, T_i, c^s_i, c^o_i) \in R : \\
            T_{i+1} \ge T_i \; \forall i \in \{1, \dots, b-1\}
            \end{array}
        \Big\} \\
        H(R) \coloneqq \Big\{ & (S_i, r_i, O_i, T_i, c^s_i, c^o_i) : (S_i, r_i, O_i, T_i, c^s_i, c^o_i) \in R, \;  i=b+1 \Big\}
    \end{aligned}
\end{equation*}
To each rule, a confidence is assigned in order to rank their predictive power relative to each other. This is done by calculating the inverse of the ratio of the number of times the rule body can be grounded in the data to the number of times each of the body groundings are followed by a fact satisfying rule head. Let the set of all possible groundings of the rule body of $R$ in TKG $\mathcal{G}$ be
\begin{equation*}
        \begin{aligned}
            &M(R, \mathcal{G}) \coloneqq \Bigg\{(s_i, r_i, o_i, t_i, c^s_i, c^o_i)_{i=1}^{b} : \\
            &\begin{array}{l} 
                (s_i, r_i, o_i, t_i, c^s_i, c^o_i)_{i=1}^{b} \models (S_i, r_i, O_i, T_i, c^s_i, c^o_i)_{i=1}^{b} \in B(R), \\ (s_i, r_i, o_i, t_i, c^s_i, c^o_i) \in \mathcal{G} \; \forall i \in \{1, \dots, b\}
            \end{array}\Bigg\}
        \end{aligned}
\end{equation*}

\noindent where $(s_i, r_i, o_i, t_i, c^s_i, c^o_i)_{i=1}^{b} \models  (S_i, r_i, O_i, T_i, c^s_i, c^o_i)_{i=1}^{b}$ indicates that the relations $\{r_i\}_{i=1}^{b}$ and categories $\{c^s_i\}_{i=1}^{b}$, $\{c^o_i\}_{i=1}^{b}$ in the left sequence equates those in the right sequence and the sequence of instantiated subjects and objects, $\{s_i\}_{i=1}^{b}$ and $\{o_i\}_{i=1}^{b}$, fulfill the order in which placeholders for the entities appear as denoted by $\{S_i\}_{i=1}^{b}$ and $\{O_i\}_{i=1}^{b}$.
The number of paths to examine when finding all possible rule groundings increases exponentially to the number of steps $b$. With a large training dataset, it becomes infeasable to find the entire set of $M(R, \mathcal{G})$. The confidence of a rule is therefore estimated by attempting to ground the rule body in the training data $n \in \mathbb{N}$ times and calculating the confidence accordingly.

Let $g_j \in M(R, \mathcal{G})$ be the $j$:th successful grounding of the rule body in $\mathcal{G}$ where $j \in J \subseteq \mathbb{N}$ such that $\text{max}(J) \leq n$. Let $o_1^{g_j}$ be the object of the first fact in $g_j$ (in terms of sequence) and let $t_b^{g_j}$ and $s_b^{g_j}$ be the timestamp and subject the $b$:th (last, in terms of sequence) fact in $g_j$, respectively. The confidence of the rule is calculated as 
\begin{equation*}
\resizebox{\linewidth}{!}{%
    $\begin{aligned}
        \label{eq:conf}
        \text{conf}(R) \coloneqq \frac{\left|\left\{\, g_j : j \in 
        J \text{ and } \exists (s, r, o, t, c^s, c^o) \in \mathcal{G} : \; t > t_b^{g_j}, \; o = s_b^{g_j}, \; s = o_1^{g_j} \right\} \right|}{\left|\left\{\, g_j : j \in 
        J \right\} \right|},
    \end{aligned}$%
}
\end{equation*}
where the denominator is the rule body support.
All learned rules are elements of a rule bank $\mathcal{B}$. Let the subset rule bank of only rules of length $b$ be $\mathcal{B}^b$. The notation $[b]$ thus constitutes all the trained rule lengths from one to $b$.

\subsubsection{Rule application}
\label{sec:rule_application}

This section describes the prediction and scoring method of C-TLogic, expanded based on TLogic. When applying the rules, the aim is to accurately predict either the object or subject of a future fact, $(s_q, r_q, ?, t_q, c^s_q, ?)$ or $(o_q, r^{-1}_q, ?, t_q, c^o_q, ?)$. For a given query, we extract a subset of rules from the rule bank. For the C-TLogic rule format $\mathcal{B}_{r_q,c^s_q} \subseteq \mathcal{B}$ consist of the rules where the head relation and subject type matches that of the query. Furthermore, let $\mathcal{G}_{t_{\text{min}}, t_q} \subseteq \mathcal{G}$ be the subset of the TKG with each timestamp $t$ within a time window of size $w$ such that $t \in [ t_q - w, t_q ) = [ t_{\text{min}}, t_q )$. The rules in $\mathcal{B}_{r_q, c^s_q}$ are applied in decreasing confidence by finding all possible groundings of the rule bodies in $\mathcal{G}_{t_{\text{min}}, t_q}$. Each body grounding results in a prediction as both entities in a rule head appears in the rule body, see Section \ref{sec:rule_learning}. As a rule can be grounded multiple times, each rule $R$ can provide a set of multiple candidate predictions. Let a candidate prediction be $k$. Each unique candidate is assigned a score by a scoring function $f:\mathcal{B}_{r_q, c^s_q} \times \mathcal{E} \rightarrow [0,1]$ and saved as $(k,f(R, k)) \in \mathcal{K}$. The scoring function $f$ is defined as:  
\begin{equation}
\label{eq:scorefunc}
    f(R, k)  
= \alpha \cdot \mathrm{conf}(R) 
+(1 - \alpha) \cdot \text{exp} ( -\lambda (t_q - \tau) ).
\end{equation}
where $\tau = \text{max}\{t_1^{g} \; \forall g \in M(R, \mathcal{G}_{t_{\text{min}}, t_q}): s_b^g=k\}$, i.e. the latest timestamp (closest to the query timestamp) of the first fact (in terms of sequence) in all the groundings of rule $R$ in subgraph $\mathcal{G}_{t_{\text{min}}, t_q}$ that predict candidate $k$. For clarification on $t_1^{g}$ and $s_b^g$, see explanation of confidence estimation in Section $\eqref{sec:rule_learning}$. The function \eqref{eq:scorefunc} minimizes the detraction of the time decay term when scoring candidate $k$. The variables $\alpha \in [0,1]$ and $\lambda > 0$ are hyperparameters. 
Each rule is applied until a chosen number ($\gamma$) of candidates with a unique number of scores are found. If $\rho$ rules have been applied so far, the requirement to stop the application of rules is thus $| \{ ( f(R_1,k), \dots , f(R_\rho,k) )^\downarrow \mid \exists i : (k, f(R_i,k)) \in \mathcal{K}\}| \geq \gamma$, where $\downarrow$ indicates sorting the scores per candidate in decreasing order.

\subsubsection{Candidate aggregation}
\label{sec:cand_ranking}

For each query we will find multiple candidates from various rules where the same candidate can be predicted by multiple rules, and scores for each candidate across rules can be aggregated by Noisy-OR and Max+ operations. 

\textbf{Noisy-OR}:
This method is applied in the original TLogic framework \cite{tlogic}. For any given candidate with related score $(k, s) \in \mathcal{K}$, the final aggregate score $s^k_{\text{final}}$ is calculated as:
\begin{equation}
\label{eq:noisyor}
  s^{(k)}_{\text{final}} = 1 - \prod\nolimits_{\{\,s \mid (k,s)\in \mathcal{K}\,\}} (1 - s).
\end{equation}
It is assumed that a candidate is correct if at least one rule presents that candidate as its prediction and each rule predicting a given candidate is assumed to be independent of any other rule predicting the same candidate. The calculation is made to aggregate this by computing the complement of all candidates failing to predict that entity. This way, even low-scored predictions can in aggregation contribute to a higher aggregated probability if many predictions of the same candidate is made. If many high scores are aggregated, the final prediction score may round off to a probability of one in implementation. This is especially a risk when the number of available options to predict is small, such as categories. In these cases, the number of predictions serve as a second ranking to decide the final ranking.

\textbf{Max+}:
As an alternative to the original Noisy-OR aggregation, we apply Max+ aggregation introduced by \cite{max+}. The candidates are ordered lexicographically, i.e. for a given candidate, the highest score of any rule predicting it is chosen as the final score and if two candidates have the same top score, the second-highest score of the two candidates is compared, proceeding sequentially until a distinction is achieved. This method differs from the Noisy-OR method in that independence between rules is not assumed. Instead, we rely solely on the highest scores. As a result, Max+ mitigates the impact of low-confidence rules.

When applying rules in entity category prediction, each prediction of a specific entity is inferred as a prediction of the entity category of that prediction. Based on this, to find the final score for a category, the two aggregations listed above are used across all predictions within the given category.

\section{Implementation}
\label{sec:implementation}
The section is organized into three main subsections. The Learner module outlines the generation of rules through temporal random walks, including the adapted sampling strategy for relation–category combinations. The Retriever module then describes how the learned rules are applied to answer queries, detailing the retrieval of relevant rules. Finally, the Evaluation module explains how these predictions are assessed against actual data, inluding how multiple correct answers are handled to ensure fair evaluation.

\subsection{Learning}
The Learner module handles the sampling of random walks, candidate rule instantiation, confidence estimation, and saving of the resulting rules. The method is, once again, similar to that of TLogic \cite{tlogic}.
As we want to keep the number of rules comparable to the TLogic framework, we make sure $n > |\mathcal{C}|$ and for each relation-category combination, we sample $\lceil \frac{n}{|\mathcal{C}|} \rceil + 1$ by breaking when $i > \lceil \frac{n}{|\mathcal{C}|} \rceil + 1$, where $i \in [n]$. This rounding up of number of walks is motivated by favoring the more complex C-TLogic. 
%
For each step of the temporal walks, facts are sampled, until reaching $b+1$ facts. 
We discard walks that are not fully completed due to lack of facts available for any step. 
In the final step $b$, the last fact is sampled from filtered facts that fulfill the cyclicality of \eqref{eq:rand_walk}, and we add the rule to the bank.
For each rule, the confidence score is computed by attempting to sample $\varsigma$ rule body groundings after which each of these rule groundings are examined if they are followed by an adequate rule head, as defined in~\eqref{eq:conf}. 
To assure the quality of the rules, the ones with confidence score above 0.01 and body support of 2 are kept (See Algorithm~\ref{alg:temporal_rule_learning}). 


The algorithm iterates over all relations $r \in \mathcal{R}$, all rule lengths $[b]$, $n$ random walks and $\varsigma$ confidence estimates. Together, this contributes with a factor of $|\mathcal{R}| b n \varsigma$. At each step in the random walk, all possible facts to traverse have to be considered, adding a factor of the maximum node degree in the training data, let it be $D$. Finally, the algorithm iterates over all $c \in \mathcal{C}$ but breaks where $i > \lceil \frac{n}{|\mathcal{C}|} \rceil$. 
So, the total worst case time complexity is $\mathcal{O}\left( | \mathcal{R}| b n \varsigma D |\mathcal{C}| \left( \frac{n}{|\mathcal{C}|} + 1 \right)\right)$.

\begin{algorithm}[t]
\caption{C-TLogic Rule Learning}
\label{alg:temporal_rule_learning}
\begin{algorithmic}[1]
  \REQUIRE Temporal knowledge graph $\mathcal{G}$
  \PARAMETERS{Maximum rule length $b_{\text{max}} \in \mathbb{N}$, number of random walks $n \in \mathbb{N}$, number of confidence samples $\varsigma \in \mathbb{N}$}
  \ENSURE Type-aware temporal logic rules  $\hat{\mathcal{B}}$
  \FOR {each relation $r \in \mathcal{R}$} 
    \FOR {each $b$ in $[b_{\text{max}}]$} 
      \FOR {$i$ in $[n]$ and $c \in \mathcal{C}$} 
          \STATE Sample a temporal random walk $W$ of length $b$ and create a rule $R$
          \FOR {$\_$ in $[\varsigma]$}
            \STATE Sample a rule body grounding and head
          \ENDFOR
          \STATE Estimating rule confidence based on sampling
          \STATE $\mathcal{B}^b_{r, c} \gets \mathcal{B}^b_{r, c} \cup \{ \langle R, \operatorname{conf}(R) \rangle \}$
        
        \IF{$i > \lceil \frac{n}{|\mathcal{C}|} \rceil $}
            \STATE \textbf{break}
          \ENDIF
      \ENDFOR
      \STATE $\mathcal{B}^b_{r} \gets \mathcal{B}^b_{r} \cup \mathcal{B}^b_{r, c}$
    \ENDFOR
    \STATE $\mathcal{B}_{r} \gets \mathcal{B}_{r} \cup \mathcal{B}^b_{r}$
  \ENDFOR
  \STATE $\mathcal{B} \gets \mathcal{B} \cup \mathcal{B}_{r}$
  \RETURN $\mathcal{B}$
\end{algorithmic}
\end{algorithm}

\subsection{Retriever}
The Retriever module is responsible for, given a query, applying the rules to the data and generating a list of candidate predictions with associated final score. 
For each query, all relevant rules are retrieved 
which share same head relation as well as subject category as the query. 
Each rule's body is grounded to data within a historical time window $w$, relative to the query timestamp. If the rule body can be grounded, a prediction can be made since cyclicality entails that the head object $o_{b+1}$ appears in the rule body as $s_b$, as denoted in~\eqref{eq:rand_walk} and \eqref{eq:second_rule}. 
Each candidate is assigned a score based on the rule confidence and the difference between the query timestamp and the rule groundings as in~\eqref{eq:scorefunc}. 
The scores are aggregated according to Noisy-OR and Max+ for each candidate predicted by multiple rules. 
The total time complexity can be computed and simplified to $\mathcal{O}\left(|\mathcal{G}| + |\mathcal{B}_{r_q, c^s_q}||\mathcal{E}| D^{b}  \right).$

\subsection{Evaluation}
The Evaluation module coordinates the queries made to the retriever, aggregates scores for each candidate, and evaluates the final rank against actual data. In cases where the retriever can not return any candidates, either due to lack of rules or inability to ground any rules, the baseline prediction is applied as described in \cite{tlogic}.

To rank and evaluate candidate entities for a given relation we use Mean Reciprocal Rank (MRR), the average of the inverse ranks, and Hits@$k$ which indicates the distribution of the ranks of correct predictions within a threshold, as it measures the proportion of times the first correct prediction appears at maximally rank position $k$, where $k$ most often take the value 1, 3, 5 and 10. Their formal definitions are as below:
\begin{equation}
    \text{MRR} = 1/|Q| \sum\nolimits_{q \in Q} (\text{rank}_q)^{-1},
\end{equation}
\noindent where $|Q|$ is the total number of queries and $\text{rank}_q$ denotes the rank position of the first correct answer for query $q$, and
\begin{equation}
\text{Hits@}k = 1/|Q| \sum\nolimits_{q \in Q} \mathbf{1}\{\text{rank}_q \leq k\},
\end{equation}
\noindent where $\mathbf{1}\{\cdot\}$ is the indicator function.



Note that as the degree of the nodes in a TKG may be larger than one, there may be multiple correct predictions, as to why the metric relies on the highest ranked correct candidate and not necessarily the specific prediction in the query.

\section{Results and Analysis}
\label{sec:res_and_an}

This chapter presents the results from the training and evaluations of the different models. 

In terms of variable selections, $n$ is chosen to be set to 200 for training on the ICEWS18 data. The FinDKG data consists of much less unique relations compared to the ICEWS18 data. When a relation more frequently appears, there are likely more patterns available to predict it. In addition, the probability that a random walk will find a valuable pattern in the data should be increasing with the number of walks. The decision of how many walks to use when training on the FinDKG data is made to equate the same number of walks per average relation frequency as in the training set of ICEWS18. With $n=200$ and an average frequency of all relations in the training set of ~1486, we get $\frac{200}{1486} \approx 0.13$. With an average frequency of all relations in the FinDKG training set of ca 14849, we choose the number of walks to be $14849 * 0.13 \approx 2000$. Based on initial experiments and the number of queries to which the rules are unable to produce a prediction, we choose to explore if increasing the number of walks will decrease the number of candidates where no prediction is possible to make for ICEWS18\_100 and ICEWS18\_12. We therefore also train the C-TLogic rules with 2000 and 500 random walks, respectively. 

Due to memory limitation, the window size for candidates is chosen $w=100$. The values $\varsigma=500$, $\alpha=0.5$, $\lambda=0.1$ are chosen following the choices of TLogic \cite{tlogic}. The number of unique predictions made before aggregating the final rankings is increased to $\gamma=30$ (from 20 in the original TLogic work) to allow for lower ranked predictions to impact final rankings, especially in category predictions.

We evaluate our method in predicting entities category of objects given query tuples. The results are shown in Table~\ref{tab:cat_perf}.
The ratio of queries where no candidates can be predicted (for example due to that no rule can be successfully grounded) is computed and denoted by "N.C.". In such cases, a baseline prediction is applied based on the entity distribution in the training data, as suggested in~\cite{tlogic}.

Note that when applying the TLogic models, categories are not taken into account and the results are thus the same across all ICEWS18-based and FinDKG-based datasets in entity-specific prediction. In addition, the original ICEWS18 dataset does not include any categories as to why C-TLogic can not be applied. 

A limitation of the evaluation is that no fine‐tuning of the models has been made due to time and computational limitations. As such, any conclusions must be understood in light of this constraint. Nevertheless, the evaluation results offer valuable insight into the relative performance of the models.

When predicting the categories, we perform an aggregation on retrieved entities that have same category (See \ref{sec:cand_ranking}). 
This way, we can apply the TLogic framework to make predictions on the entity categories. 
Generally, the TLogic model performs stronger than C-TLogic on the ICEWS18 datasets. Within TLogic, the Max+ aggregation again outperforms Noisy-OR aggregation. This indicates that its more valuable looking at the highest ranked predictions per category rather than including low-scoring inferences. By contrast, C-TLogic performs stronger with Noisy-OR aggregation, suggesting that the confidence of these rules does not as accurately reflect the actual predictive power of the rules on the test set. Overall, TLogic surpasses C-TLogic regardless of aggregation choice. On the FinDKG dataset, the Noisy-OR aggregation yields a very high superior performance in both categorization settings.

When comparing classification granularity, performance declines as the number of categories increases. This suggests a trade‐off between fine‐grained categorization and the ease with which rules can generalize across clusters. Additionally, an imperfect clustering or GICS sector‐mapping might also degrade performance.
%
%
%
\begin{table*}[htbp]
    \centering
    \begin{threeparttable}  
    \begin{subtable}[t]{\textwidth}
        \centering
        {\small
            \begin{tabular}{ll|lccccc|lccccc}
                \toprule
                \multicolumn{2}{c}{\textbf{Model}}
                 & \multicolumn{6}{c}{\textbf{ICEWS18\_12}}
                 & \multicolumn{6}{c}{\textbf{ICEWS18\_100}} \\
                \cmidrule(lr){1-2}\cmidrule(lr){3-8}\cmidrule(lr){9-14}
                Rule Format & Cand.\ Agg.
                 & Walks & N.C. & \textbf{MRR} & \textbf{H@1} & \textbf{H@3} & \textbf{H@10}
                 & Walks & N.C. & \textbf{MRR} & \textbf{H@1} & \textbf{H@3} & \textbf{H@10} \\
                \midrule
                    TLogic & Noisy‑Or & 
                        200 & 4.02 & 75.79 & 68.21 & 79.71 & 93.64 & 
                        200 & 4.02 & 43.10 & 34.19 & 44.85 & 62.35  \\
                    TLogic & Max+ &   
                        200 & 4.02 & 79.53 & 72.22 & 84.81 & 93.91 & 
                        200 & 4.02 & 60.38 & 53.01 & 63.35 & 75.36   \\
                    C-TLogic & Noisy‑Or &
                        500 & 8.65 & 60.48 & 50.36 & 70.27 & 75.82 & 
                        2000 & 20.50 & 31.44 & 25.22 & 36.53 & 41.94 \\
                    C-TLogic & Max+ &     
                        500 & 8.65 & 57.87 & 46.18 & 69.11 & 75.82 &  
                        2000 & 20.50 & 27.95 & 20.15 & 34.26 & 41.91  \\
                \bottomrule
            \end{tabular}

        }
        \label{tab:entity_perf-a}

    \end{subtable}

    \begin{subtable}[t]{\textwidth}
        \centering
        {\small
            \begin{tabular}{ll|lccccc|lccccc}
                \toprule
                \multicolumn{2}{c}{\textbf{Model}}
                 & \multicolumn{6}{c}{\textbf{FinDKG}}
                 & \multicolumn{6}{c}{\textbf{FinDKG\_Comp}} \\
                \cmidrule(lr){1-2}\cmidrule(lr){3-8}\cmidrule(lr){9-14}
                Rule Format & Cand.\ Agg.
                 & Walks & N.C. & \textbf{MRR} & \textbf{H@1} & \textbf{H@3} & \textbf{H@10}
                 & Walks & N.C. & \textbf{MRR} & \textbf{H@1} & \textbf{H@3} & \textbf{H@10} \\
                \midrule
                   TLogic & Noisy‑Or & 
                        2000 & 1.99 & 92.44 & 90.45 & 93.23 & 97.52 & 
                        2000 & 1.99 & 89.84 & 87.55 & 90.48 & 95.54 \\
                    TLogic & Max+ &   
                        2000 & 1.99 & 66.87 & 52.88 & 75.72 & 97.28 & 
                        2000 & 1.99 & 64.68 & 51.50 & 72.69 & 94.51 \\
                    C-TLogic & Noisy‑Or &
                        2000 & 3.47 & 65.24 & 53.99 & 74.09 & 86.08 & 
                        2000 & 4.89 & 59.48 & 48.04 & 68.83 & 80.13 \\
                    C-TLogic & Max+ &     
                        2000 & 3.47 & 61.35 & 47.04 & 73.44 & 86.12 & 
                        2000 & 4.89 & 56.89 & 43.39 & 68.39 & 80.13 \\
                \bottomrule
            \end{tabular}
        }

    \end{subtable}
    \caption{Performance of category prediction models. N.C., \textbf{MRR}, \textbf{H@1}, \textbf{H@3} and \textbf{H@10} are denoted in percentages.}
    \label{tab:cat_perf}
    \end{threeparttable}
        
  \vspace{-15pt} 
\end{table*}
\section{Discussion}
Based on the experiments, the choice of rule format in combination with aggregation method has a pronounced effect on predictive performance. 
Despite a higher number of random walks, the higher complexity of the C-TLogic rules seems not to be able to capture patterns TLogic is able to find and apply generally as the ratio of queries where no prediction can be made is high. This may be a key factor in the accuracy difference and potentially suggests the method of learning the rules either has to be more dynamic to address these gaps or be more relaxed in application.

The reason why the C-TLogic does not reach the performance of the original TLogic framework can be a bias-variance issue. Including the entity categories in the rule format introduce major differences compared to TLogic. Firstly, in training, the rules are trained to predict a given subject category-relation combination and sequences of facts that do not involve this combination in the head fact are excluded. This exclusion does not allow for any valuable general pattern for prediction across categorizations to be captured in favor of the remaining subject category-relation pairs. Consequently, this limits rule application to only those rules that has captured a pattern for the given subject category-relation combination in the training set. A general patterns across categories may have been overlooked or may not even exist in the dataset for the given subject category-relation combination.

\begin{figure}[htbp]
  \centering

  \begin{subfigure}[t]{\linewidth}
    \centering
    \includegraphics[width=\linewidth]{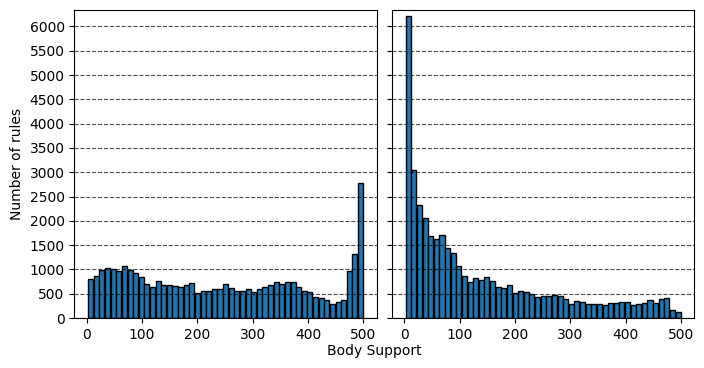}
  \end{subfigure}

  \begin{subfigure}[t]{\linewidth}
    \centering
    \includegraphics[width=\linewidth]{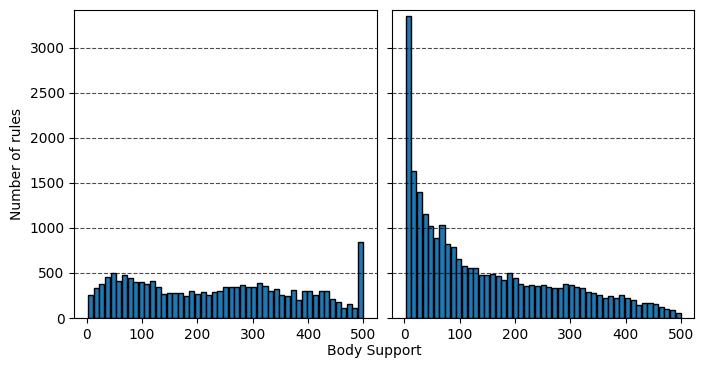}
  \end{subfigure}
  \caption{Frequency of rules per body support level for rules trained on the datasets
  ICEWS18\_12 (top) and FinDKG\_Comp (bottom), when training TLogic (left) and C-TLogic (right).}
  \label{fig:body_support_per_rule_all}
  \vspace{-15pt} 
\end{figure}

The high accuracy of Max+ in combination with TLogic does motivate further investigation into the aggregation methods used in the later parts of the prediction task. In the category predictions, the superior performance of TLogic is similar to the entity predictions. In C-TLogic, the rule formats are constructed to capture some more information regarding how categories interact compared to TLogic, following the hypothesis of this paper. The results indicate that the potential benefit attributable to capturing these patterns is less than the benefit from finding and applying general patterns across categories. Still, we can show the TLogic favorably can be applied for the new purpose of category-prediction and reach a very high accuracy on datasets such as FinDKG.

The clustering of the entities was motivated by allowing rule application to be facilitated based on entity category. Intuitively, the clustering can provide a logical categorization to help understand the distribution of entities. There are multiple potential explanations for why this necessarily does not result in a higher prediction accuracy with the C-TLogic rule format. One likely reason is that there are patterns that are general and apply well across entity categories that the limitation of training with a set number of walks per relation-category combination misses. Moreover, a pattern captured by one walk predicting a certain relation-subject category combination may be valueable for a second relation-subject category combination with the same relation, but this pattern may not even exist in the training data for the second relation-subject category combination. The TLogic takes advantage of the shared pattern by allowing application across categories which C-TLogic does not.
Another possible explanation relates to the higher ratio of queries to which no prediction can be made with C-TLogic models. A potential solution here was to increase the number of walks which did improve the ratio in the case of ICEWS18\_12 but not to a satisfactory level (for the C-TLogic with 200 walks resulting in a N.C. ratio of 11.59\% decreased to 8.65\% with 500 walks). Naturally, the number of possible combinations of relations and entity category pairs increases greatly and finding a large proportion of the valuable patterns in this more granular data requires a much larger number of walks. It also adds to the issue discussed above with overlooking general patterns across categories. Lastly, the relative degradation of performance on the FinDKG\_Comp dataset underscores a trade‐off between granularity in categorization and predictive power. Although suboptimal sector mapping cannot be ruled out, the drop suggests that overly coarse categorization inherently limits benefits of category‐aware rule formats.



\section{Conclusions and Future work}
\label{sec:conclusionsAndFutureWork}


This work has investigated TKG link prediction at both entity and category levels. A new rule format, C-TLogic, has been introduced based on the original TLogic framework, with the purpose of finding category specific patterns and facilitating more relevant rule application. In addition, an alternative aggregation method, Max+, has been incorporated into both TLogic and C-TLogic. To ensure comparability on the ICEWS18 dataset, entities were categorized using clustering based on the semantic similarity of entity labels. The number of clusters was chosen by optimizing the Bayesian Information Criterion and by matching the category count of the FinDKG dataset.
In this paper a viable procedure for categorizing entities was presented- whether using existing taxonomy or via semantic clustering - to enrich TKGs. By grouping entities into meaningful categories, one can tailor inference to the level of granularity useful for downstream decision-making tasks. Examples range from sector-specific risk assessment in finance to event-type forecasting in political analysis. 

The adjustments proposed in this paper show that the original TLogic framework can be effectively used in category prediction, extending its use-case applicability. In practice, the goal is not often to predict a specific entity with a lower level of accuracy but instead use predictive methods complementary to support conclusions or other work. Extending the FinDKG categories based on the GICS sectors will produce a very high level of accuracy, both using the original categories and the expanded categorization. This indicates that the performance cost when increasing specificity is not too high. Letting the use-case steer the categorization of the dataset could be a viable way to use this model. In addition, the alternative aggregation method Max+ can be used to increase the accuracy in some cases. 
How to optimally choose aggregation method and data pre-processing remains a relevant area for future research. 

Another path forward is to look into a more applied context where you let the need of the application steer the categorization and thus attempt to develop rules for a specific application purpose. An example is the moat and tailwind stock classification within equity asset management that describes how different companies answer differently to changes in markets and economic conditions.

The methods explored in this paper along with much other research aims to make predictions at a certain point in time (the timestamp of the query). In the confidence estimation section of the TLogic framework the time distance between the rule grounding and the head which it is predicting is not taken into account. Incorporating this information into the estimation and subsequently in rule application could improve prediction ranking. 

\bibliographystyle{ACM-Reference-Format}
\bibliography{References}

\end{document}